\documentclass[letterpaper, 10 pt, conference]{ieeeconf}  % Comment this line out
\IEEEoverridecommandlockouts                              % This command is only
\usepackage{graphicx}
\DeclareGraphicsExtensions{.pdf,.png,.jpg}

\title{\LARGE \bf
Anomaly Detection in Multivariate Non-stationary Time Series for Automatic DBMS Diagnosis
}

\author{ \parbox{3 in}{\centering Doyup Lee*
         \thanks{Data, workstation and efficacy of experiment results are produced by EXEM Co.}\\
         Department of Creative IT Engineering\\
         Pohang University of Science and Technology\\
         77 Cheongam-ro Nam-gu, Pohang, Gyeongbuk, Republic of Korea\\
         zzehqlzz@postech.ac.kr}
}

\begin{document}

\maketitle
\thispagestyle{empty}
\pagestyle{empty}

%%%%%%%%%%%%%%%%%%%%%%%%%%%%%%%%%%%%%%%%%%%%%%%%%%%%%%%%%%%%%%%%%%%%%%%%%%%%%%%%
\begin{abstract}

Anomaly detection in database management systems (DBMSs) is difficult because of increasing number of statistics (stat) and event metrics in big data system. In this paper, I propose an automatic DBMS diagnosis system that detects anomaly periods with abnormal DB stat metrics and finds causal events in the periods. Reconstruction error from deep autoencoder and statistical process control approach are applied to detect time period with anomalies. Related events are found using time series similarity measures between events and abnormal stat metrics. After training deep autoencoder with DBMS metric data, efficacy of anomaly detection is investigated from other DBMSs containing anomalies. Experiment results show effectiveness of proposed model, especially, batch temporal normalization layer. Proposed model is used for publishing automatic DBMS diagnosis reports in order to determine DBMS configuration and SQL tuning. \\

\end{abstract}


%%%%%%%%%%%%%%%%%%%%%%%%%%%%%%%%%%%%%%%%%%%%%%%%%%%%%%%%%%%%%%%%%%%%%%%%%%%%%%%%
\section{INTRODUCTION}

Database management systems (DBMSs) become the critical part of Big Data application because of hardness to control increasing number of data and its scale [17]. Anomaly detection in DBMS is important to 1) prevent from crucial problems such as DBMS black out and low performance, and 2) determine configuration and SQL tuning point from database administers (DBAs). Previously, DBAs have investigated anomaly periods with DBMS disoders and found causal events manually based on their own experience. However, control of modern DBMS has surpass human ability as the size and complexity of DBMS grow [1].

"Anomaly" can be defined in different way in respect from data characteristics ans its domain [14]. In practical view, I indicate anomaly in time series as unpredictable or unexplainable change which can't be observed and inferred. For example, sudden additive outliers or trend changes are anomalies in unpredictability perspective. Unpredictability approach is appropriate to define anomaly in practical manner. For example, it is not crucial if the changes were predictable in past, although some changes arise.

Anomaly detection is hard to implement with machine learning because of expensive labeled data and few number of data [5]. In addition, anomaly detection is important in professional fields such DBMSs, IT securities, or industrial monitoring [14]. Eventually, making labeled anomaly data needs difficult processes by experts in each domain. However, it is impossible with expensive cost for experts to construct labeled anomaly dataset finding and recording few number of anomalies every time. Thus, many previous anomaly detection models can't be utilized in practice because most of the models are based on supervised learning approach [11, 16]. Especially, in time series cases, previous researches have limited to only stationary or periodic time series data.

With the development of big data technology, deep learning have outperformed in various area such as image recognition, natural language process [18, 21]. Deep learning models enable to extract hidden representation in big data which human can't find easily with hand-craft programming. Among them, deep autoencoder finds hidden representation in high dimensional data with multi layers encoder-decoder structure, extracting lower dimensional abstract features to reconstruct of input data [19].

In this paper, I propose anomaly detection model based on reconstruction error using autoencoder with non-stationary time series data in DBMSs. Proposed model aims to detect anomalies in time series DBMS metrics, determine time periods containing anomalies, and find causal events related with anomalies in the period.\\

\section{Theoretical Background}
\subsection{Anomaly Detection}
Anomaly detection refers to finding unexpected behaviors that do not conform to observed pattern [5]. In machine learning, it means data patterns that are not available in training or insufficient to train because of few number of samples. Based on model characteristic, anomaly detection models are divided into 5 categories: probabilistic, distance-based, reconstruction-based, domain-based, and information-theoretic model. Especially, neural network-based model can be utilize to detect anomalies as reconstruction-based model [14]. 

Reconstruction-based anomaly detection models use reconstruction error as criteria of anomaly or anomaly score. That is, it assorts data patterns into anomaly if the patterns have high error score after reconstruction. As reconstruction models, neural network-based models containing encoder-decoder structure and subspace-based model such as (kernel) PCA are commonly used [2, 6, 15]. Setting error threshold is one of the problem to utilize reconstruction-based model for anomaly detection.

\subsection{Deep Autoencoder}

An autoencoder is unsupervised learning model that has encoder-decoder structure with feed-forward neural network. It trains reconstruction model which infers input $x$ as output. Input $x$ is encoded into feature space $C(x)$ and decoded into original space. In general, autoencoder is used to extract hidden representation of input distribution with lower dimensionality [7]. In this paper, autoencoder is trained and used in order to reconstruct input data. With multi-layers neural network in encoder and decoder, deep (stacked) autoencoder can extract more abstract and complex representation of input distribution [19].

\subsection{Statistical Process Control}
Statistical Process Control (SPC) is a statistical model to monitor outputs of a process. The objective of SPC is to find assignable causes that are unpredictable and lower process capability or performance. In order to know whether assignable causes exist or not, control charts are used in SPC. Control charts monitor characteristic value of a process such as sample standard deviation and sample mean of process output. 

When sample mean is used, it is called $\overline{X}$ chart. The average of sample mean, $\overline{\overline{X}}$ is set as center line of control chart. In addition, $\overline{\overline{X}} \pm 3\sigma$ become upper control limit (UCL) and lower control limit (LCL), where $\sigma$ is standard deviation of sample means [12]. 

\begin{figure}
\centering
\includegraphics[scale=.3]{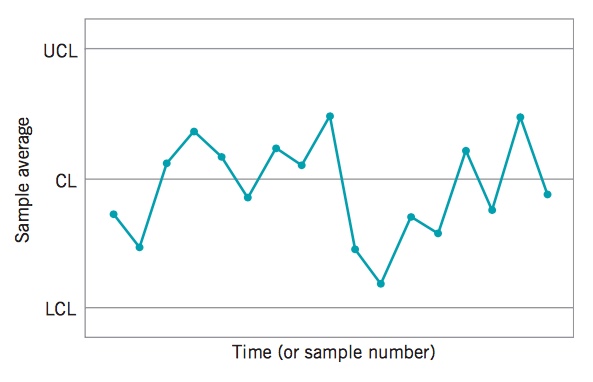}
\caption{Control Chart Example [13]}
\end{figure}

A process is consider out-of-control when assignable causes exist in the process. There are many rules for identifying out-of-control and the most typical rule is to decide it when characteristic value $\overline{X}$ is located outside of control limits, UCL or LCL. When it occurs, the process is considered containing assignable causes that have to be eliminated in order to keep high stability and capability of the process.

Previous reconstruction-based anomaly detection models have a limitation on how to determine appropriate anomaly threshold in different data [11, 15]. However, when it is combined with SPC, anomaly threshold can be set based on statistical background on process control aspect.

\subsection{Non-stationary Time Series}
Let $\textbf{Z} ={\left\{Z_t\right\}}_{-\infty}^{+\infty}$ is a sequence of random variables, called time series. It is said to be strictly stationary if its distribution is invariant to time periods. 
$$
F_{Z_{t_1} ,..., Z_{t_n}}(x_1, ..., x_n) = F_{Z_{t_1+k} ,..., Z_{t_n+k}}(x_1, ..., x_n) \eqno{(1)}
$$

Stationary time series is hard to define in real world data, so weak stationarity condition is usually applied. Weak stationarity means first and second moments are invariant to time periods. 

In contrast, non-stationary time series has different distribution according to time periods. In other words, non-stationary time series has time-varying means and variance. For example, time series which has trend of time has non-stationarity. In time series analysis, non-stationary time series is converted into stationary series using de-trending, differencing, and other data transformation techniques. [20]

When a machine learning model treats time series data, stationarity is also important in respect with function approximation. If input data has non-stationarity, the model tries to learn different input distribution as input changes and eventually fails to learn successfully. In summary, converting non-stationary to stationary time series is essential when machine learning model learns its parameters.\\

\section{Proposed Model}

\subsection{Problem Description}
To define problem, I interviewed senior DBAs and DB consultants who work in DBMS monitoring company in Republic of Korea. The company has most of market share (above 70 \%) of DB performance management market in Korea. 

When some defects and disorders occur in database, DBMS users call DB consultants in order to find causal time periods and events. After finding, consultants resolve the problems, changing DBMS configurations or tune wrong SQL. However, from detecting disorder periods to taking solutions, consultants have to depend on their own past experience and manual efforts when they work. They spend most of time to find disorder periods and causal events in log data analysis process, because there are hundreds of event metrics in DBMS. 

In order to analysis DBMS log data, they manually find time point that main DB stats and wait events suddenly change. After then, they confirm anomaly periods and find wait events, similar with change patterns of main stat metrics. Based on causal wait events detected, they implement solutions like configuration or SQL tuning. 

Although DBMS configuration and SQL tuning might need professional knowledge of DBAs and consultants, detection of causal periods and related events can be found automatically by DBMS metrics. It reduces consultants' time consumption enough. The objective of proposed model is automatic diagnosis that finds time period with anomalies and causal events related with DB anomalies.

\subsection{Data}

\begin{figure}
\centering
\includegraphics[scale=.12]{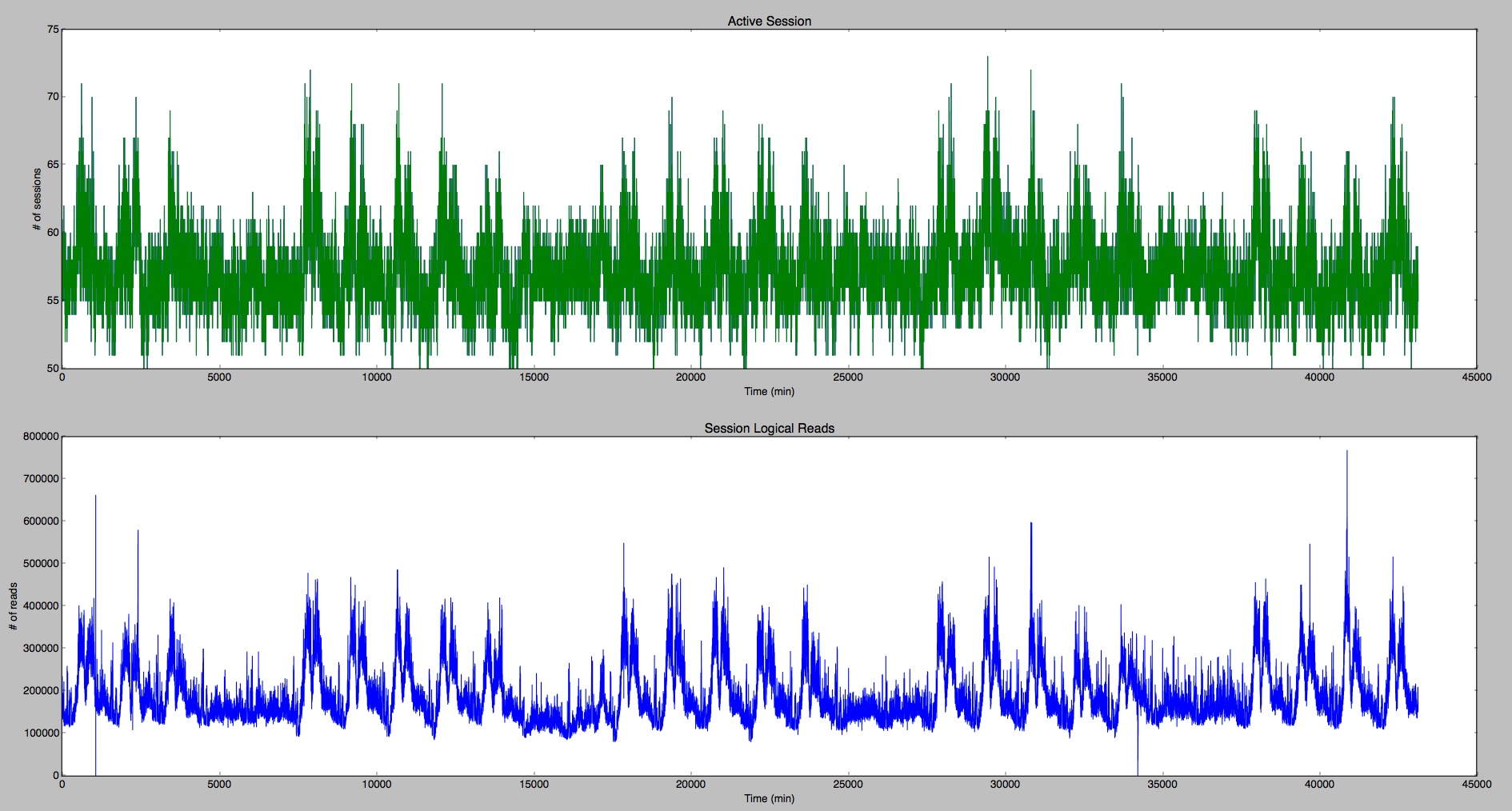}
\caption{Stat Metrics Example: Active Session and Session Logical Reads}
\end{figure}

In this paper, two kinds of time series data are used: 1) DBMS Statistics (stat) Metrics and 2) DBMS Events Metrics. Data is collected in a minute. 

\subsubsection{DBMS Stat Metrics}
Stat metrics describe basic information of DB status that how DB works. There are hundreds of stat metrics, but 6 number of main stat metrics are used in this paper. The stat metrics are selected based on DBAs and consultants' interview. The selected metrics are main ones that DB consultants consider when find anomaly periods. Example plots of stat metrics data are in Figure 2, active session and session logical reads in minutes.
\begin{itemize}
\item CPU Used:
How much CPU is used when it collected.
\item Active Session: 
The number of sessions that executes SQL.
\item Session Logical Reads: 
The number of requested blocks for execution of queries
\item Physical Reads: 
The number of blocks that disk reads.
\item Execute Counts:
The number of execution of SQL
\item Lock Waiting Session:
The number of sessions that waits in lock.\\
\end{itemize}

\subsubsection{DBMS Event Metrics}
DBMS event metrics describe wait information that occurs in DB when sessions request queries. For example, direct path read, SQL net message from client, and LGWR wait for redo copy are contains in event metrics. There are also hundreds of event metrics. They are used to diagnose causes of DB disorders at DBMS configuration and SQL levels in anomaly periods.

\subsection{Model Scheme}

\begin{figure}
\centering
\includegraphics[scale=.3]{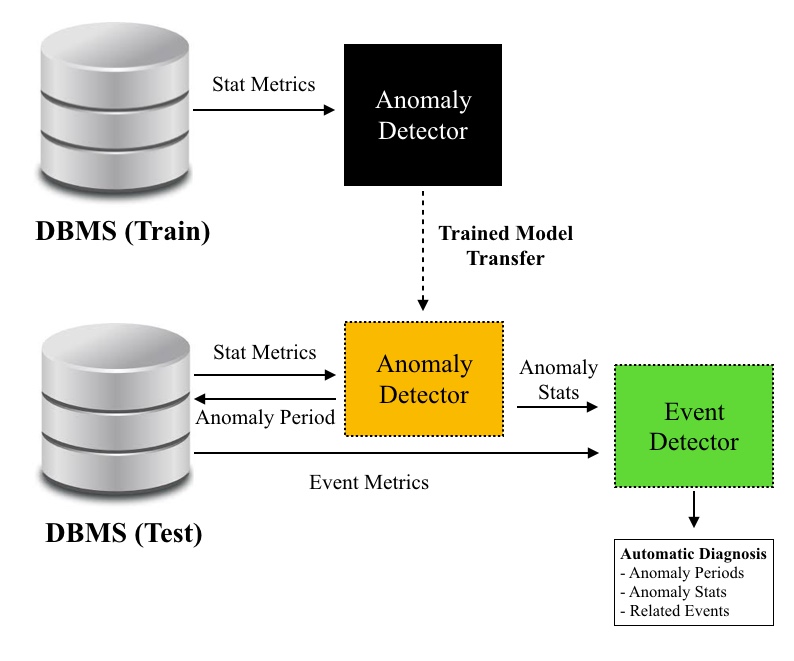}
\caption{Proposed Model Scheme}
\end{figure}

Figure 3. shows model scheme that is proposed for detecting anomalies in this paper. From DBMS stat metrics data, anomaly detector that determine time periods with anomalies is trained. After training model, the model is utilized to detect anomaly periods in other DBMSs, not same DBMS. In other words, trained model is transfered to other DBMSs and additional training is optional in here. 

Production of pre-trained model is important in service providers' perspective. There are many customer companies that want to detect anomalies automatically before consultation of their DBMSs. If there is no pre-trained model, each company, even each DBMS, can't uses automatic anomaly diagnosis until they store their DBMS metrics and train their machine learning model with optimal hyper-parameters. After pre-trained model is produced with basic and valid performance, each DBMS can be customized by additional training process after they store their data. Thus, in this paper, efficacy of transfered model is verified after training anomaly detector.

When anomaly detector finds anomaly periods, causal events that have most similar pattern with anomaly stat metric are investigated based on time series similarity measures such as dynamic time warping (DTW) [4, 9] or Pearson correlation [3].

Finally, anomaly periods, related stat, and event metrics are reported automatically to DB users and consultants. Based on diagnosis result, consultants change DBMS configuration or specific SQL in order to solve and prevent from DB disorders.

\subsection{Anomaly Detector Model Description}
\begin{figure}
\centering
\includegraphics[scale=.2]{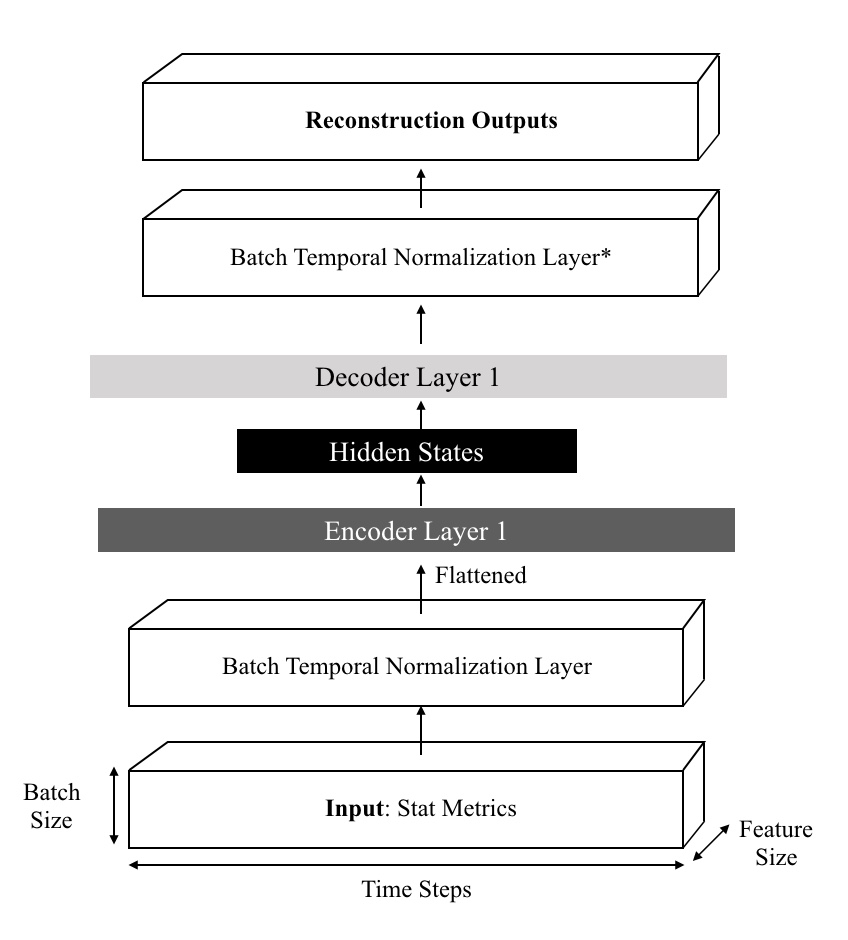}
\caption{Anomaly Detector Model Architecture}
\end{figure}

In order to detect anomalies in DBMSs, reconstruction approach is used in proposed model. Deep autoencoder trains DBMS stat metrics data and it determines anomaly periods based on reconstruction error. In other words, output that has high reconstruction error is interpreted as anomaly sample which can't be predictable with patterns of past stat metrics. 

Figure 4 shows deep learning model architecture for anomaly detection. With repeated experiments, specific architecture is set with 3 hidden layers autoencoder. In this model, time series inputs are connected to batch temporal normalization layer in order to confirm stationarity of each sample. In decoder part, reverse calculation is proceeded corresponding to encoder part.

\subsection{Batch Temporal Normalization}
Batch normalization is critical technique for fast learning speed and generalization [8]. In this paper, batch temporal normalization layer is proposed for stationarity of input time series. It normalizes input data when batch input data is selected as batch normalization does. However, it does not have same calculation with batch normalization because the objective is not correspondence of feature scale, but correspondence of moments in different time periods. 

Figure 5 shows operational difference between batch normalization and batch temporal normalization. Yellow part is the calculation range of each normalization operation. Batch temporal normalization calculates mean and standard deviation of each sample on time steps, but batch normalization calculates them both on batches and time steps. Batch temporal normalization makes each sample have same mean and standard deviation regardless of time periods for stationarity of input. I note that this calculation is essential when machine learning model treats non-stationary time series data. Same as with batch normalization, scale and offset parameters are trained in batch temporal normalization layer.

\subsection{Anomaly Period Detection with SPC}
In test part of Figure 3, pre-trained anomaly detector is used to determine time periods containing anomalies. Proposed model uses SPC approach based on reconstruction error. In order words, sample anomaly score in (2) is a characteristic value of control chart that monitors process stability. If sample anomaly score is out of control limits, corresponding time period is considered anomaly period when related events need to be found.

$$
{(Sample\quad Anomaly\quad Score)}_{ij} = \frac{1}{T}\sum_{t=1}^{T}({\hat{y}}_{i,j,t} - y_{i,j,t})^2 \eqno{(2)}
$$
where i,j is sample and feature index. 

\begin{figure}
\centering
\includegraphics[scale=.25]{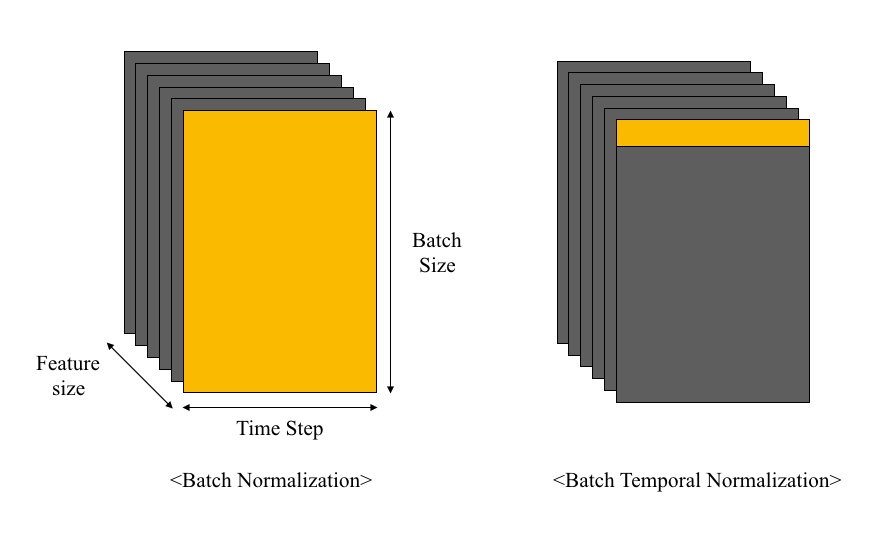}
\caption{Difference between Batch Normalization and Batch Temporal Normalization}
\end{figure}

\section{Experiments}

\subsection{Training Results of Deep Autoencoder}
In order to find appropriate hyper-parameters of deep autoencoder model before training, repetitive experiments were proceeded. The number of layers and neurons, optimizer, learning rate, regularization parameter, and batch size were selected after empirical repetitions. ADAM optimizer [10], 0.001 learning rate and regularization parameter, and 1500 size of batch are used to train the model. Proposed model is implemented using Python 2.7.12, Tensorflow 1.0.1, and Numpy 1.12.1. Workstation with a Titan X GPU is used to train proposed model.

For training, 6 stat metrics described above section are collected from a DBMS in a minute for a month. Training time series data is constructed with time window 30 steps in order to detect anomaly periods in 30 minutes. Collected data is divided into 60\%, 20\%, and 20\% for training, validation, and test. Trained model with lowest validation loss is selected as final trained model for model generalization. Activation function in all layer except output is Rectified Linear Unit (ReLU). ReLU is proper to real value time series than sigmoid. Mean square error is used to train model as loss function. For faster learning, input features are normalized with moments as to total time period. 

$$
Train\quad Loss = \frac{1}{N}\sum_{i=1}^{N}\sum_{j=1}^{F} \sum_{t=1}^{T}({\hat{y}}_{i,j,t} - y_{i,j,t})^2 \eqno{(4)}
$$

Table 1 shows test results of various deep autoencoder models. Asterisks in Table 1 mean reverse calculations in decoder part of autoencoder. I found that batch temporal normalization improves performance of deep learning model that input time series has non-stationarity. In Table 1, simple (150)-(50)-(150*) model has lowest test error although it doesn't include batch temporal normalization layer. However, in next subsection, it was not selected as optimal anomaly detector because is has no ability to detect anomaly periods. Although models with BTN have higher test loss than simple one, they could detect anomaly periods accurately. Finally, BTN-(150)-(50)-(150*)-(50*)-BTN* model was selected as optimal anomaly detector. I note that batch temporal normalization layer is essential to take internal dynamics of non-stationary time series in deep learning model. Figure 6 shows learning curve of optimal anomaly detector in log scale.

\begin{table}[h]

\label{table_example}
\begin{center}
\begin{tabular}{|c|c|c|c|}
\hline
Model (Neurons) & Test (MSE)\\
\hline
PCA-network (50) & 0.7951\\
\hline
PCA-network (50) with BTN & 1.1560\\
\hline
(150)-(50)-(150*) & \textbf{0.3096}\\
\hline
(150)-BN-(50)-BN*-(150*) & 0.8620\\ 
\hline
BN-(150)-(50)-(150*)-BN* & 1.2432\\
\hline
BTN-(150)-(50)-(150*)-BTN* & \textbf{0.6912}\\
\hline
BTN-(150)-BN-(50)-BN*-(150*)-BTN* & 0.7803\\
\hline
BN-(500)-(300)-(150)-(300*)-(500*)-BN* & 4.4816 \\ 
\hline
BTN-(500)-(300)-(150)-(300*)-(500*)-BTN* & 0.7034 \\ 
\hline
BTN-(500)-BN-(300)-(150)-(300*)-BN*-(500*)-BTN* & 0.6925 \\ 
\hline
\end{tabular}
\end{center}
\caption{Test Result of Trained Anomaly Detector Models}
\end{table}

\begin{figure}
\centering
\includegraphics[scale=.13]{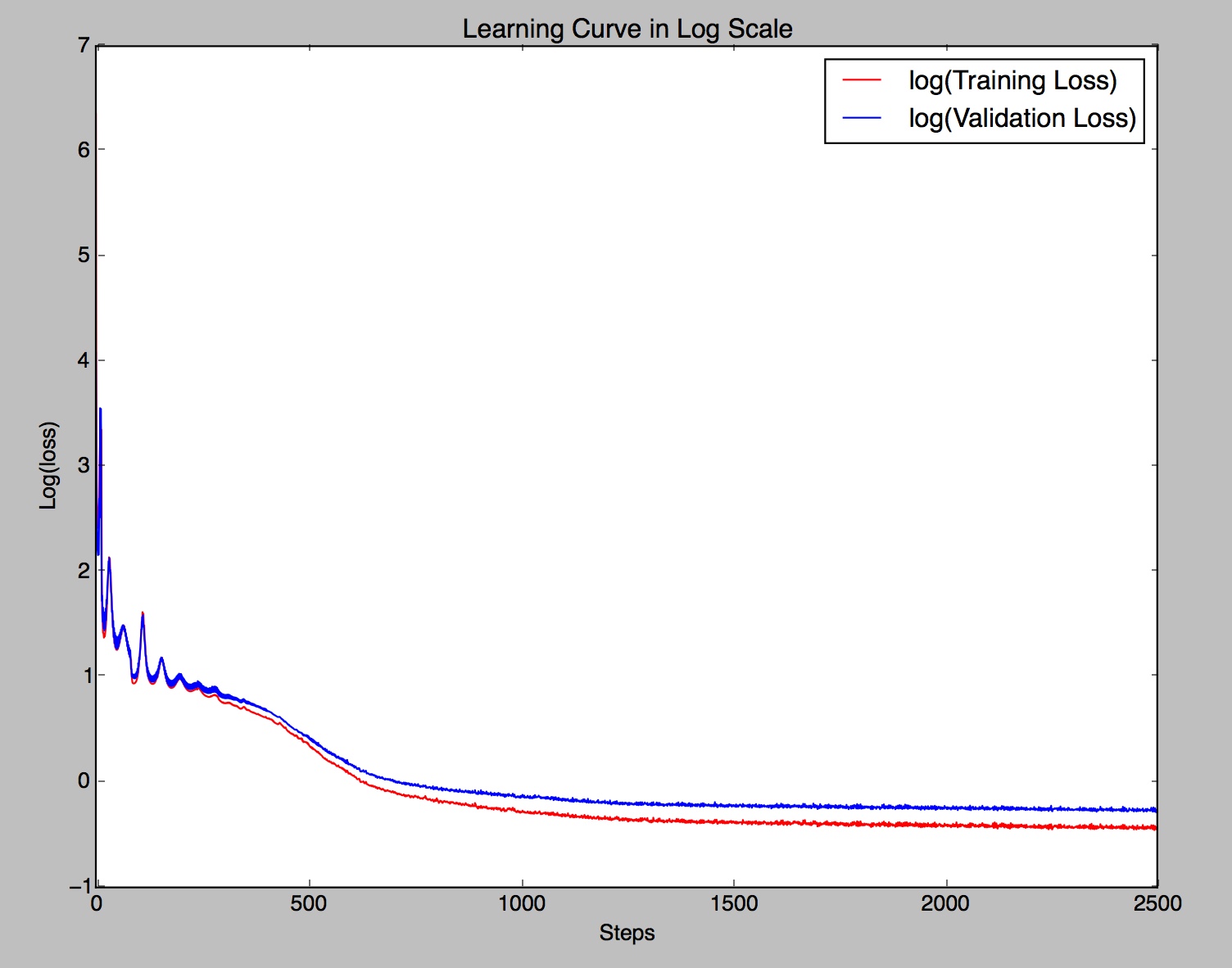}
\caption{Learning Curve of BTN-(150)-(50)-(150*)-BTN*}
\end{figure}

\subsection{Anomaly Periods Detection}
Anomaly periods detection test is proceeded in other DBMSs, where training data is not from. Detection performance was validated by blind tests on past DBMS disorder samples that DBAs had diagnosed. I validated efficacy of anomaly periods detection with 10 disorder cases. Proposed model detects multiple periods as anomalies, but DBAs recorded only one periods as disorder periods. However, top-1 error period corresponded to DBAs' opinion. I shows some example of anomaly periods detection.

\begin{figure}
\centering
\includegraphics[scale=.2]{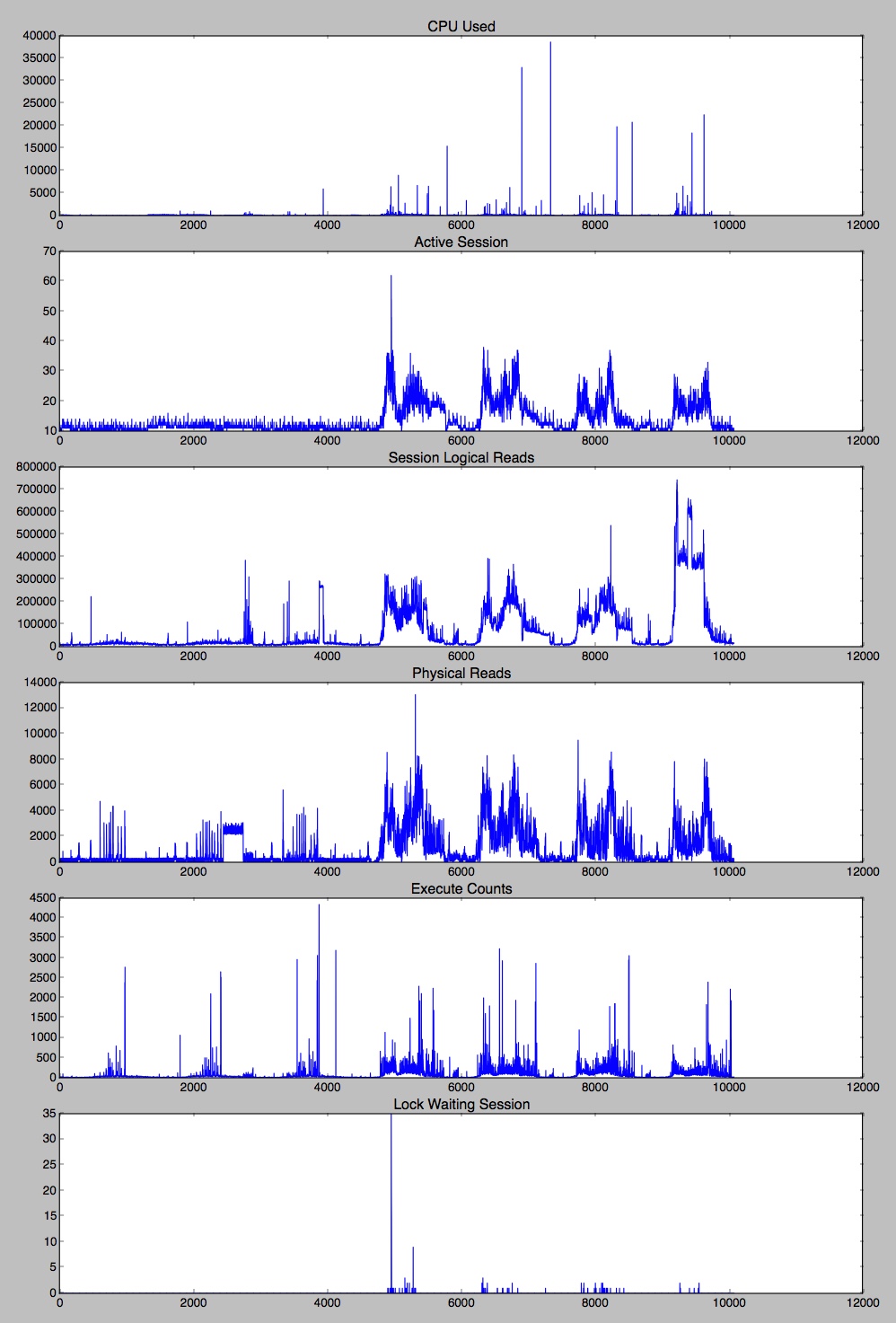}
\caption{DBMS Disorder Case 1}
\end{figure}

Figure 7 shows DBMS disorder case 1. DBAs diagnosed disorder period when active session suddenly peaked (in second graph). Figure 8 shows the result of anomaly periods detection by proposed (optimal) model. Red lines indicate upper central limit, center line, and lower central limit in SPC. It shows that anomaly detector successfully finds all time periods that each stat metric has radical changes, especially sudden peaks. Of course, it detects the time period that DBAs concluded as disorder period with much high anomaly score.

Figure 9 shows anomaly detection results of (150)-(50)-(150*) model that has lowest test loss in previous subsection. The anomaly score patterns are similar regardless of feature type and detect almost same results except few periods. Especially, the pattern is close to 'CPU used' anomaly score pattern in Figure 8. I checked that it was not a mistake on programming and the terrible results are from the absence of BTN layer. I found scales of Figure 9 are extreme high. For example, in active session case, selected model scores anomaly error up to 80 in Figure 8, but Figure 9 shows about 4000. Anomaly scores in the other features have same problem in Figure 9. 

Main causes of this problem are inferred with 3 reasons. First, all metrics in training data have similar pattern in same time. Second, detection model is trained with multivariate time series data, not univariate. It means relationship between input metrics effects on output. In addition, radical change in a feature can have several effects on other features if the model is not well-trained. Third, the trained model without BTN is seriously sensitive to specific feature value, in here CPU used. Especially, it is critical problem of the model in Figure 9, considering the data is even preprocessed before inference. Thus, reconstruction error of all metrics have extremely high value when specific metric has anomaly value. It is empirically shown that BTN prevent from high dependency on specific features with non-stationary time series data. Figure 8, 9, and 10 are another test example and its anomaly detection results. It shows same results with case 1.

\begin{figure}
\centering
\includegraphics[scale=.2]{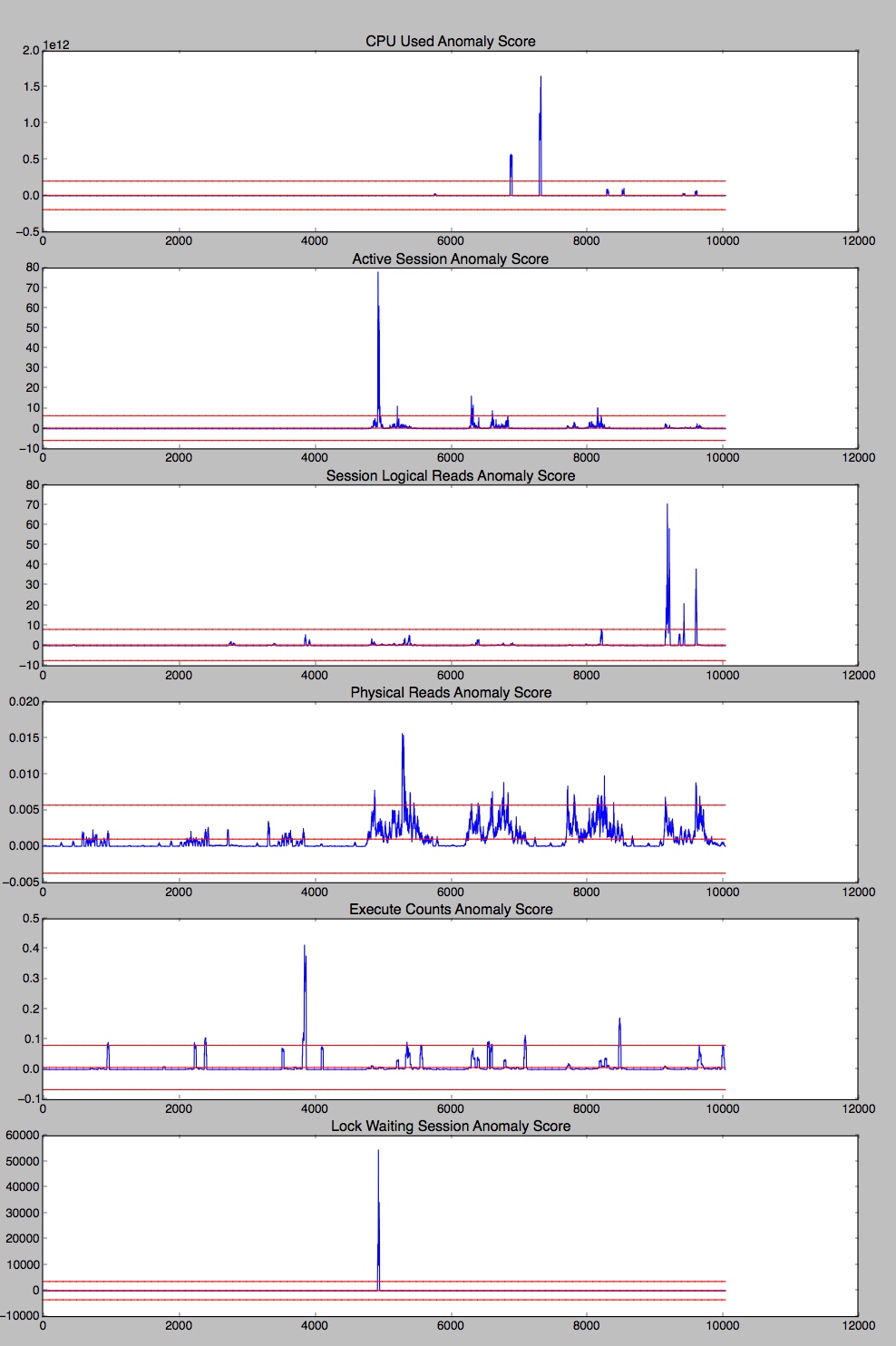}
\caption{Anomaly Detection Result of Case 1}
\end{figure}

\begin{figure}
\centering
\includegraphics[scale=.2]{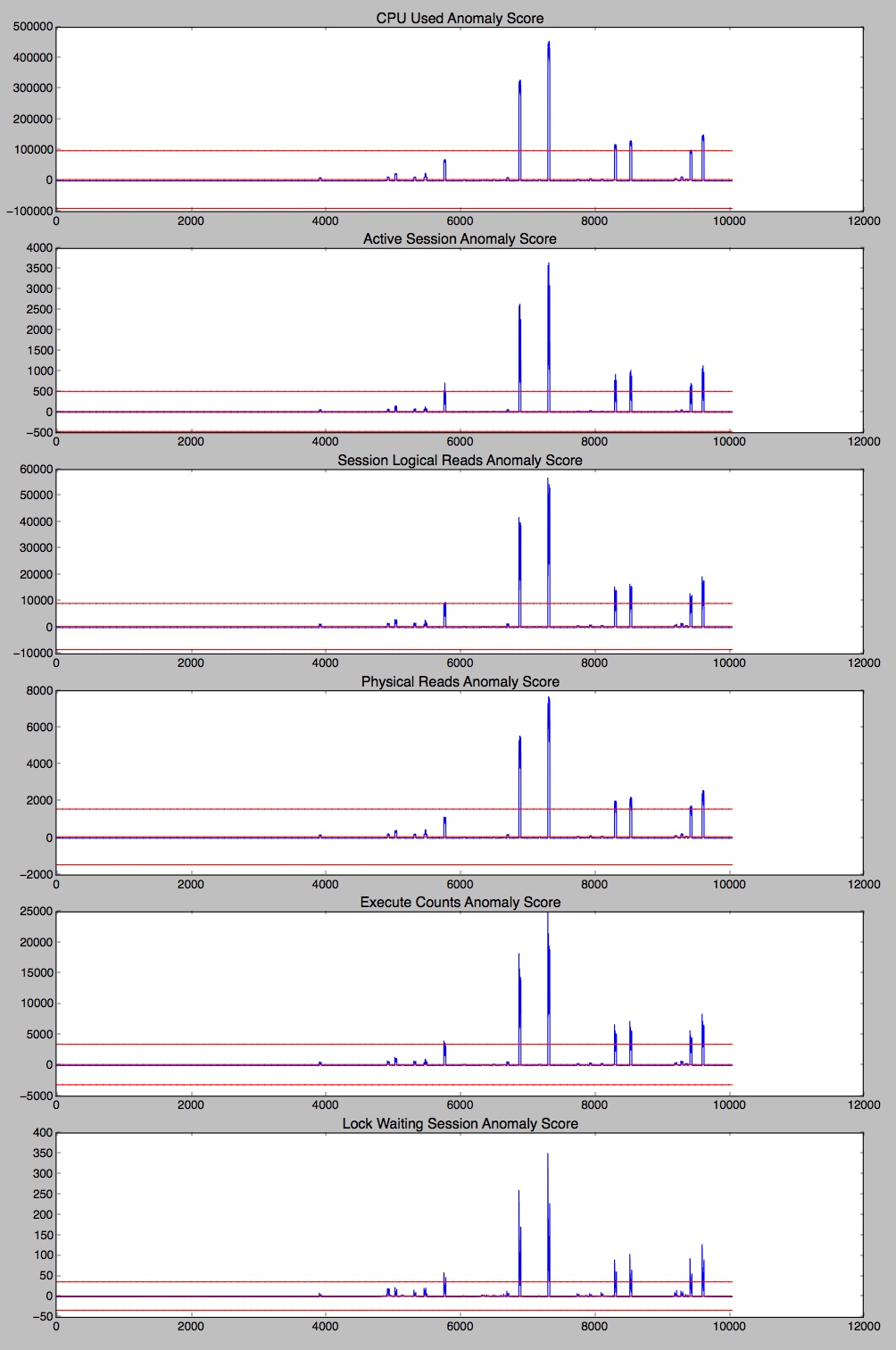}
\caption{Anomaly Detection Result of Case 1 with no BTN}
\end{figure}

Although DBAs and consultants detect an anomaly periods in a sample with their manual effort, proposed model suggests multiple candidates of anomaly period based on numerical anomaly score. If the results are provided to DBAs when they detect disorder periods, it helps them make complex results beyond their experience and save time to detect.

\begin{figure}
\centering
\includegraphics[scale=.2]{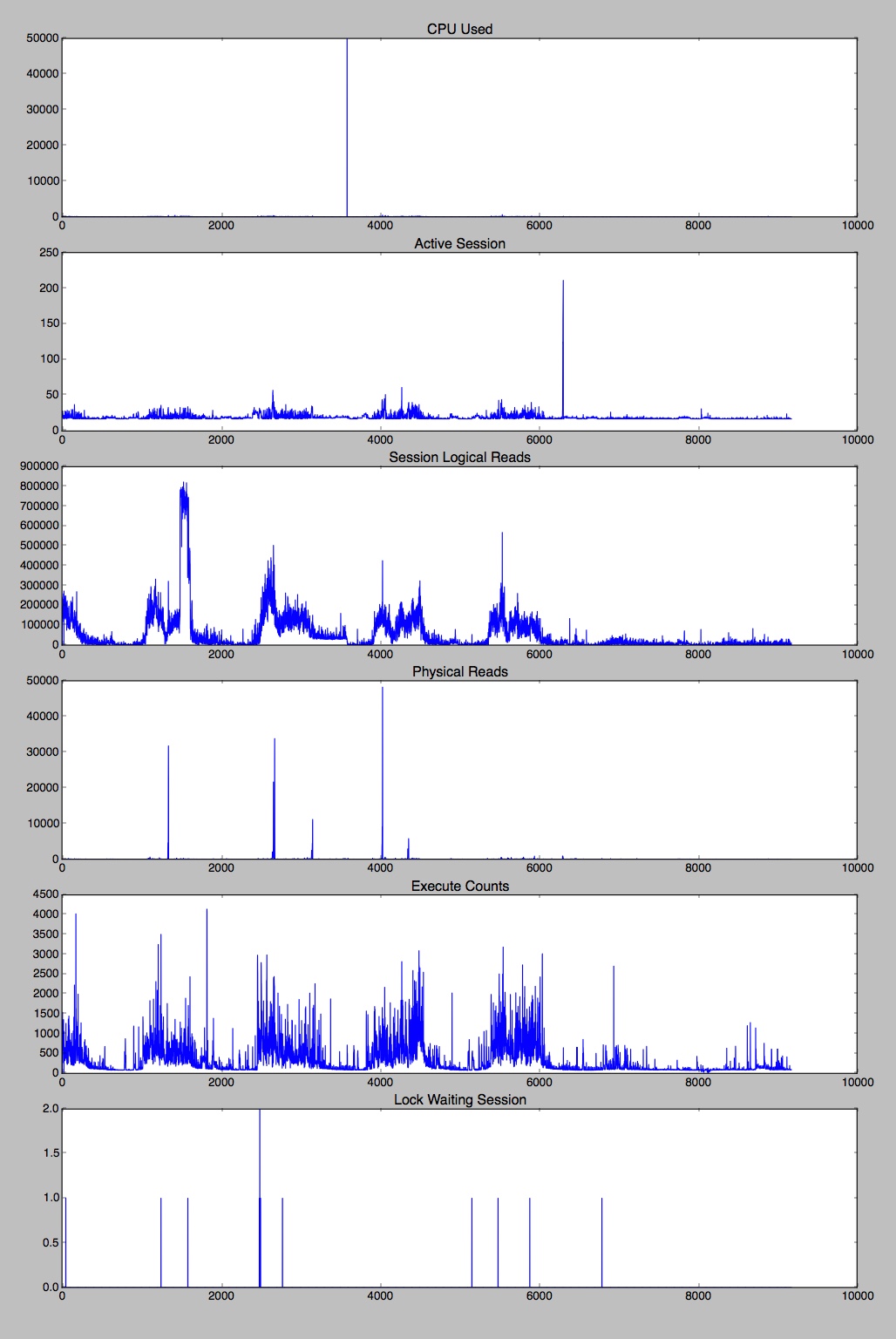}
\caption{DBMS Disorder Case 2}
\end{figure}

\begin{figure}
\centering
\includegraphics[scale=.2]{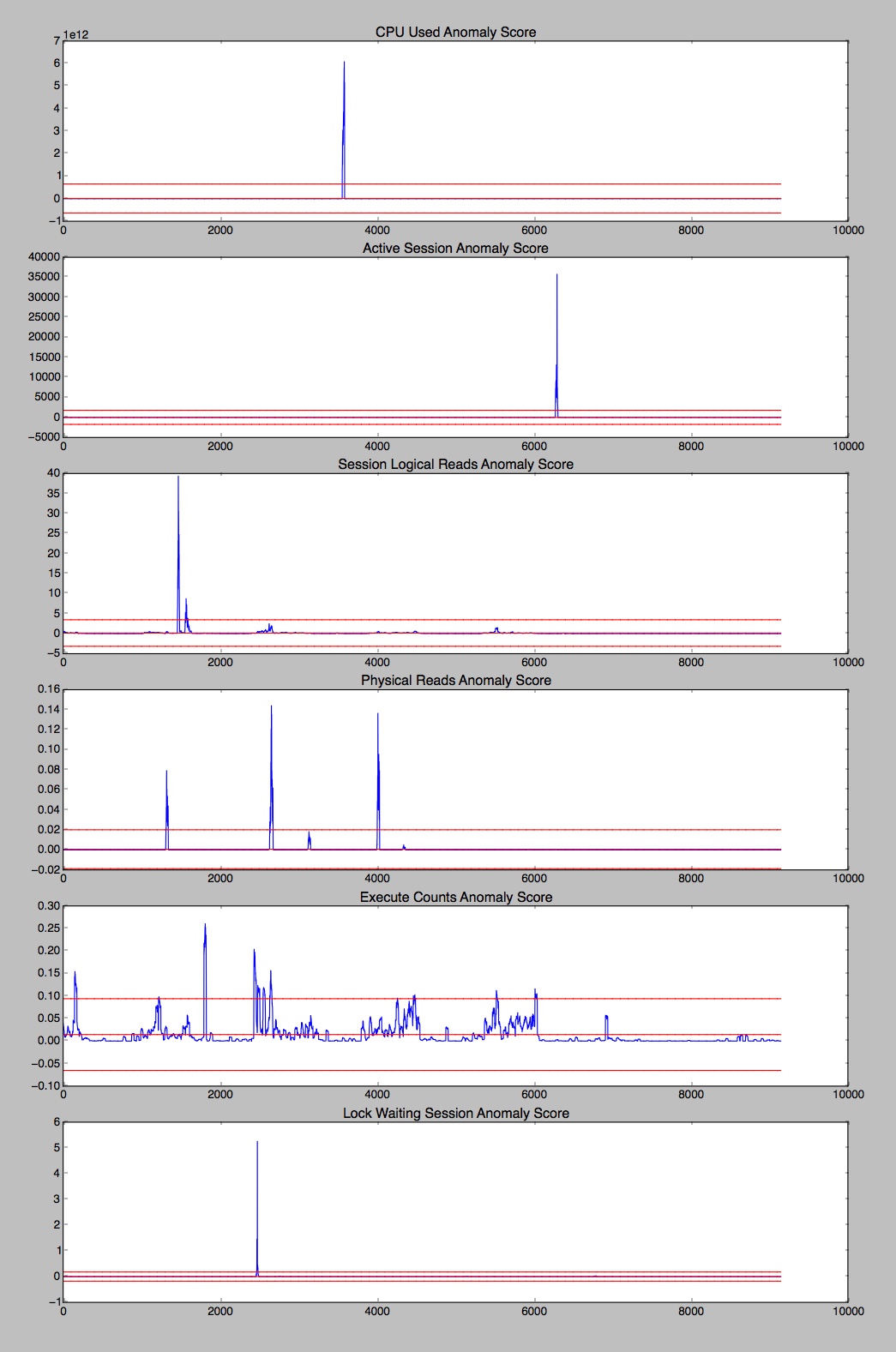}
\caption{Anomaly Detection Result of Case 2}
\end{figure}

\begin{figure}
\centering
\includegraphics[scale=.2]{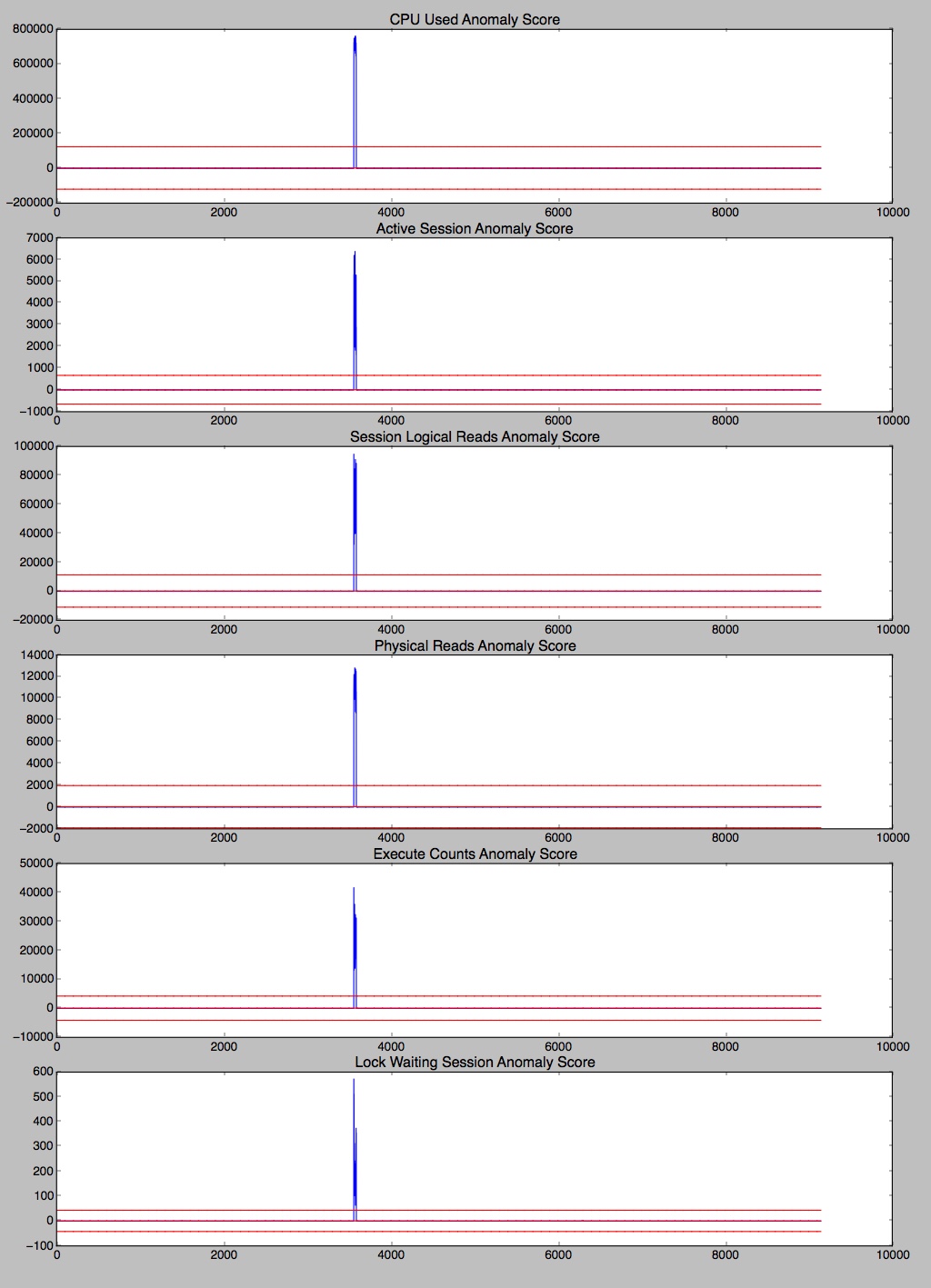}
\caption{Anomaly Detection Result of Case 2 with no BTN}
\end{figure}

\subsection{Casual Events Detection}
After detecting anomaly period samples, that is, out-of-control time period in SPC, causal events have to be found in order to figure out specific reasons of disorder. After DBAs understand main reason causing anomaly events, they take some solutions such as DBMS configuration or SQL tuning.

Time series distance/similarity measures are used to detect causal events that are most related with anomaly patterns of corresponding stat metrics in the period. Dynamic Time Warping (DTW) and Pearson correlation are both calculated in order to find best matched event with stat metric in anomaly period. In this paper, related events with anomaly period in case 1 when active session suddenly peaks is attached below. 

Figure 13 shows example of stat metrics in anomaly period and results of best matched events derived by distance/similarity measures. The difference of measures can be known in Figure 13. DTW doesn't calculate distance limited to corresponding time step pairs, but Pearson correlation only calculates with corresponding time step. DTW focuses on distance of period in real value perspective and Pearson correlation does similarity of time step in increase/decrease ratio perspective. In time step level, pattern of the event with highest Pearson correlation is similar with pattern of active sessions. However, the value of event doesn't increase dramatically. 

In contrast, overall pattern of the event with lowest DTW doesn't seem similar with active sessions, it dramatically increases right after active session suddenly peaks in extreme value. When considering that causal events don't arise in same time with stat metrics, DTW seems to be appropriate measure to detect causal events in anomaly period. However, in this model, I decided to produce both measure information because of DBAs' request.

\begin{figure}
\centering
\includegraphics[scale=.2]{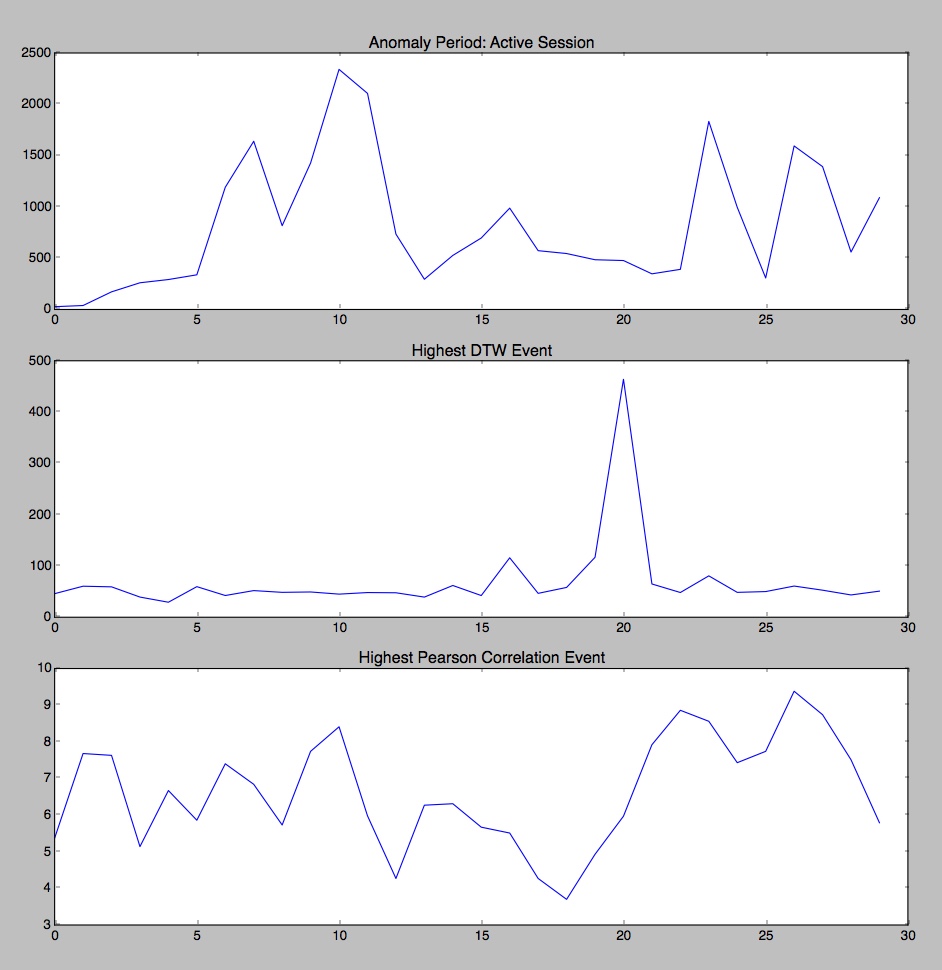}
\caption{Related Event Examples in Case 1}
\end{figure}

\section{Conclusion and Future Work}
I proposed a machine learning model for automatic DBMS diagnosis. The proposed model detects anomaly periods from reconstruct error with deep autoencoder. I also verified empirically that temporal normalization is essential when input data is non-stationary multivariate time series. 

With SPC approach, time period is considered anomaly period when reconstruction error is outside of control limit. According types or users of DBMSs, decision rules that are used in SPC can be added. For example, warning line with 2 sigma can be utilized to decide whether it is anomaly or not [12, 13]. 

In this paper, anomaly detection test is proceeded in other DBMSs whose data is not used in training, because performance of basic pre-trained model is important in service providers' perspective. Efficacy of detection performance is validated with blind test and DBAs' opinions. The result of automatic anomaly diagnosis would help DB consultants save time for anomaly periods and main wait events. Thus, they can concentrate on only making solution when DB disorders occur.

For better performance of anomaly detection, additional training can be proceeded after pre-trained model is adopted. In addition, recurrent and convolutional neural network can be used in reconstruction part to capture hidden representation of sequential and local relationship. If anomaly labeled data is generated, detection result can be analyzed with numerical performance measures. However, in practice, it is hard to secure labeled anomaly dataset according to each DBMS. Proposed model is meaningful in unsupervised anomaly detection model that doesn't need labeled data and can be generalized to other DBMSs with pre-trained model.

\addtolength{\textheight}{-12cm}   % This command serves to balance the column lengths

%%%%%%%%%%%%%%%%%%%%%%%%%%%%%%%%%%%%%%%%%%%%%%%%%%%%%%%%%%%%%%%%%%%%%%%%%%%%%%%%

%%%%%%%%%%%%%%%%%%%%%%%%%%%%%%%%%%%%%%%%%%%%%%%%%%%%%%%%%%%%%%%%%%%%%%%%%%%%%%%%

%%%%%%%%%%%%%%%%%%%%%%%%%%%%%%%%%%%%%%%%%%%%%%%%%%%%%%%%%%%%%%%%%%%%%%%%%%%%%%%%
\section*{ACKNOWLEDGMENT}

“This research was supported by the MSIT(Ministry of Science and ICT), Korea, under the ICT Consilience Creative program(IITP-2017-R0346-16-1007) supervised by the IITP(Institute for Information \& communications Technology Promotion)”

%%%%%%%%%%%%%%%%%%%%%%%%%%%%%%%%%%%%%%%%%%%%%%%%%%%%%%%%%%%%%%%%%%%%%%%%%%%%%%%%

\end{document}